\documentclass[pdflatex,sn-basic, Numbered]{sn-jnl}

\usepackage{graphicx}%
\usepackage{multirow}%
\usepackage{amsmath,amssymb,amsfonts}%
\usepackage{amsthm}%
\usepackage{mathrsfs}%
\usepackage[title]{appendix}%
\usepackage{xcolor}%
\usepackage{textcomp}%
\usepackage{manyfoot}%
\usepackage{booktabs}%
\usepackage{algorithm}%
\usepackage{algorithmicx}%
\usepackage{algpseudocode}%
\usepackage{listings}%

\usepackage{hyperref}
\usepackage{placeins}
\usepackage{multirow}
\usepackage{makecell}
\usepackage{booktabs}
\usepackage{color,soul}
\usepackage{mdframed}
\usepackage{float}
\usepackage{longtable}

\newmdenv[
  linecolor=gray!50,         
  backgroundcolor=gray!10,   
  linewidth=2pt,             
  roundcorner=5pt,           
]{graybox}

\begin{document}

\title[Article Title]{Remembering Unequally: Global and Disciplinary Bias in {LLM Reconstruction of Scholarly Coauthor Lists}}


\author*[1]{\fnm{Ghazal} \sur{Kalhor}}\email{kalhor.ghazal@ut.ac.ir}

\author[2]{\fnm{Afra} \sur{Mashhadi}}\email{mashhadi@uw.edu}

\affil*[1]{\orgdiv{School of Electrical and Computer Engineering, College of Engineering}, \orgname{University of Tehran}, \orgaddress{\city{Tehran}, \country{Iran}}}

\affil[2]{\orgdiv{Computing and Software Systems}, \orgname{University of Washington}, \orgaddress{ \city{Bothell}, \state{WA}, \country{USA}}}


\abstract{Ongoing breakthroughs in large language models (LLMs) are reshaping scholarly search and discovery interfaces. While these systems offer new possibilities for navigating scientific knowledge, they also raise concerns about fairness and representational bias rooted in the models' memorized training data. As LLMs are increasingly used to answer queries about researchers and research communities, their ability to accurately reconstruct {scholarly coauthor lists} becomes an important but underexamined issue. In this study, we investigate how memorization in LLMs affects the {reconstruction of coauthor lists} and whether this process reflects existing inequalities across academic disciplines and world regions. We evaluate three prominent models---DeepSeek R1, Llama 4 Scout, and Mixtral 8×7B---by comparing their generated {coauthor lists} against bibliographic reference data. Our analysis reveals a systematic advantage for highly cited researchers, indicating that LLM memorization disproportionately favors already visible scholars. However, this pattern is not uniform: certain disciplines, such as Clinical Medicine, and some regions, including parts of Africa, exhibit more balanced reconstruction outcomes. These findings highlight both the risks and limitations of relying on {LLM-generated relational knowledge} in scholarly discovery contexts and emphasize the need for careful auditing of memorization-driven biases in LLM-based systems.}

\keywords{Large language models, LLM memorization, Disciplinary and regional bias, {Coauthor list reconstruction}, Fairness and inclusion in scholarly discovery}



\maketitle

\section{Introduction}\label{sec1}

Over the past year, large language models (LLMs) have been increasingly integrated into research workflows, particularly for supporting literature reviews and summarizing scientific papers. Tools like Elicit AI \cite{kung2023elicit} search over 125 million articles from the Semantic Scholar database and use LLMs to generate summary tables and findings. Other frameworks, such as LitLLM~\cite{agarwal2024litllm}, ChatCite~\cite{li2024chatcite}, and LLAssist~\cite{haryanto2024llassist}, have also emerged to facilitate AI-assisted academic review. While these systems are gaining traction among researchers, there has been limited scrutiny of whether they represent scholars and publications equitably across academic disciplines and regions. {Although such systems do not explicitly retrieve coauthor lists, they rely on latent representations of authors and their relationships to generate search results, summaries, and recommendations. In this context, an LLM's ability to reconstruct an author's coauthor list serves as a diagnostic proxy for how that author and their scholarly connections are internally represented by the model. Systematic inaccuracies or biases in these reconstructions may therefore signal unequal representational capacity, which can propagate to downstream tasks in scholarly information retrieval.} 

At the same time, studies have shown that LLMs are prone to memorizing frequently encountered training data, especially when exposed to repeated instances during early training stages~\cite{Touvron2023LLaMAOA,wang2024generalization,saparov2023languagemodelsgreedyreasoners}. \citet{carlini2022quantifying}, for example, demonstrated that models with as few as 6 billion parameters can reconstruct entire passages from training sources like textbooks. More recent work~\cite{hayes2025measuring} has quantified such memorization through probabilistic querying. However, current research focuses primarily on verbatim text reproduction. There is little understanding of how memorization might affect relational outputs, such as inferred {coauthor lists}, or whether such effects are consistent across domains and geographies. This gap motivates our investigation into how LLMs differentially encode scholarly relationships and whether such encoding introduces representational disparities.

To investigate this, we examine whether LLMs memorize {coauthor lists} differentially as a function of citation frequency. {LLMs are primarily trained on the full text of scholarly articles, which includes not only the main content but also their reference lists.} {Although LLMs also encounter author lists in the headers of individual papers, these occurrences are comparatively sparse relative to references, which repeatedly reproduce the same bibliographic entries across many documents.} As a result, highly cited papers---and the bibliographic entries associated with them---appear repeatedly across the training corpus. {This repeated exposure increases the frequency with which the names of highly cited researchers and their collaborators occur in the data seen by the model, making such author lists more amenable to memorization. Consequently, LLMs are more likely to recall and reconstruct the coauthor lists of highly cited researchers with greater accuracy and consistency.} Based on this reasoning, we hypothesize that citation frequency systematically shapes LLM memorization: researchers whose names appear more often in bibliographies are more likely to be accurately represented in LLM-generated {coauthor lists}, independent of discipline or region. This mechanism implies that memorization-driven outputs may disadvantage lesser-cited researchers and reinforce existing disparities in scholarly visibility.


To test this hypothesis, we critically examine the reliability of the LLM-generated {coauthor lists} by comparing them with OpenAlex and Google Scholar, two {large-scale bibliographic databases with broad disciplinary and geographic coverage.} For {1,596} seed authors selected across 10 academic disciplines and 8 global regions, we collect their {coauthor lists} using data from {OpenAlex} and Google Scholar. We then query three widely used LLMs, differing in model size and potentially trained on data from various global regions where access to open scientific literature may vary. The LLMs are DeepSeek R1 (671 billion parameters), Llama 4 Scout (17 billion parameters), and Mixtral 8x7B (7 billion parameters). We use these LLMs in context-free prompting to reconstruct {coauthor lists}. We systematically evaluate discrepancies across various academic disciplines and geographical regions to answer the following research questions:

\begin{itemize}
    \item \textbf{RQ1:} Can we observe the effect of memorization in the {coauthor lists} generated by LLMs? To what extent does this memorization increase as a function of model size and {author citation count (used as a proxy for training data repetition)?}
    \item \textbf{RQ2:} Is the effect of memorization evenly distributed across academic disciplines?
    \item \textbf{RQ3:} Does the influence of memorization show a consistent pattern across different regions of the world?
\end{itemize}

Our findings provide clear evidence of memorization in LLMs, with significantly higher Discoverable {Name} Extraction (DNE) scores for highly cited authors compared to their lower-cited counterparts across all models. {We introduce the DNE score as a novel metric to measure how well an LLM reproduces real-world {coauthor lists} when prompted with a single author's name} (see Subsection~\ref{subsec:memorizationMetric}). This indicates that highly cited researchers are more accurately and frequently represented in LLM-generated {coauthor lists}. We also observe that memorization effects are more pronounced in models with a greater number of parameters. While these effects are broadly present, we identify important exceptions at the academic discipline level. In particular, Clinical Medicine shows no significant difference in DNE scores between highly and lower-cited authors, suggesting more equitable representation. At the regional level, similar patterns emerge in Sub-Saharan Africa (for DeepSeek R1) and North Africa (for Mixtral 8x7B), where the {coauthor lists} of lower-cited researchers are represented on par with those of highly cited researchers. These findings point to areas where training data may be more balanced, highlighting academic disciplines and regions that exhibit reduced representational bias in LLM outputs.

This research underscores the need for a nuanced approach to utilizing LLMs for scholarly {discovery tasks}, {for instance, in identifying or mapping scholars' {coauthor lists}}, particularly with regard to equitable global academic representation. We discuss implications for the broader adoption of LLMs in scientometric applications and propose improvements to enhance inclusivity and equal representation of scholarly work across diverse academic landscapes. Finally, we ensure the reproducibility of our research by making all data and analysis code publicly available.

\section{Related work}

\subsection{Coauthorship representation biases}

{Coauthorship networks are widely used to analyze patterns of collaboration and scholarly influence. Prior studies have shown that network characteristics, such as degree centrality and clustering, can predict research impact, including metrics like the h-index} \cite{grodzinski2021can}. {Researchers have also documented social and demographic disparities in coauthorship networks, including persistent gender imbalances within academic communities} \cite{bravo2019gender,kalhor2022new,macedo2023academic,lopez2022mirage}.

{Despite the growing interest in LLMs, only one study to date has investigated their application in reconstructing {coauthor lists} and identifying associated biases. In our previous work \cite{kalhor2025measuring}, we introduced AI-generated coauthorship networks, using LLMs to retrieve coauthors for a list of researchers. We analyzed potential biases in these networks with respect to gender and ethnicity by comparing LLM-generated results to data from Google Scholar and DBLP. Our findings suggest that while there was no evidence of gender bias in the LLM-generated names, there was clear evidence of ethnic bias, specifically, an overrepresentation of Asian and White names in the retrieved {coauthor lists}, as well as higher network accuracy for researchers belonging to these ethnic groups.

\subsection{Memorization in LLMs}

The phenomenon of memorization in LLMs refers to their propensity to encode and retrieve verbatim or near-verbatim content from their training datasets. This behavior has significant implications for model performance, factuality, and privacy. A number of studies have investigated when and how memorization occurs, noting that  for rare or low-frequency inputs, memorization is more prevalent in larger models,  \citep{wang2025generalization, bombieri2024llms, magar2022datacontaminationmemorizationexploitation}.~\citet{carlini2022quantifying} showed how models as little as 6 Billion parameters are capable of reconstructing text from their training sample, using discoverable extraction measures.~\citet{hayes2025measuring}   quantified   memorization through probabilistic querying by modifying Discoverable Extractions to  accommodate for probabilistic and non-deterministic nature of queries. 

While most prior work has focused on memorization of individual phrases or documents, relatively little is known about whether LLMs memorize more complex relational patterns, social or academic {relationships among entities}. Recent research on structured generation tasks, including knowledge graph completion and entity-relation extraction, suggests that LLMs can learn and reproduce relational data \citep{ranaldi2025protoknowledge, richardeau2024llms}. However, the fidelity and origin of these patterns, whether learned statistically or memorized directly, remain open questions.

Our work is the first to our knowledge to examine how memorization may manifest in the LLM reconstruction of {coauthor lists}. {To distinguish between statistical learning and memorization, we compare LLM-generated lists with ground-truth bibliometric data across multiple academic disciplines and geographic regions, using targeted prompts designed to elicit memorized content.} This approach is particularly relevant in light of globally uneven publication practices and disciplinary specializations. Previous studies have documented geographic and topical imbalances in LLM training data \citep{manvi2024large}, which may contribute to disproportionate memorization of certain regions or academic fields. We investigate whether such biases lead to structural disparities in LLM-generated {coauthor lists} across academic disciplines and regions, providing new insight into the sociotechnical limitations of integration of LLMs into scientometric.

\section{Methodology}
\subsection{Data collection}
In this section, we describe our methodology for collecting a balanced and unbiased dataset of seed authors and coauthors.

\subsubsection{Selection of seed authors}
To select a balanced representative sample of authors, we choose 10 fields of science, as defined in the classification system of the Stanford/Elsevier Top 2\% Scientists List 2024\footnote{\url{https://top2percentscientists.com/stanford-elsevier-top-scientists-list-2024/}}, each encompassing several subfields. The full list of fields and their corresponding subfields is provided in Table~\ref{tab:field}. From these subfields, we collect all the authors who have a Google Scholar page listing the subfield in their Google Scholar profile and have a citation count higher than 100. We choose this cut-off number to ensure that the LLMs recognize the author's publications as part of their training data. In addition to the fields, we define 8 regions of the globe, corresponding to North America, South/Central America, Europe, North Africa, Sub-Saharan Africa, Middle East, East/Southeast Asia, and Oceanic. For each author, we collect their affiliation and email domain from Google Scholar using the Scholarly API~\cite{cholewiak2021scholarly}. We then use this information to determine their country of affiliation via the Google Maps API~\cite{googlemapsapi}, which is used to assign each author to a corresponding world region.

\subsubsection{Grouping of seed authors}
For each academic discipline and each region, we randomly select 10 authors (referred to as seeds henceforth) from both the first and fourth interquartile ranges of citation counts. This process yields a total of 1,600 initial seed authors. {We assume that} seed authors with high citation counts are more likely to represent training example repetition, as their names are frequently cited alongside their coauthors. Conversely, authors with lower citation counts {may correspond to} data points that are less frequently encountered in the training datasets of LLMs. This grouping enables us to test our hypothesis on LLM memorization and to address the research questions posed.

\begin{figure*}
      \centering
      \includegraphics[width=\linewidth]{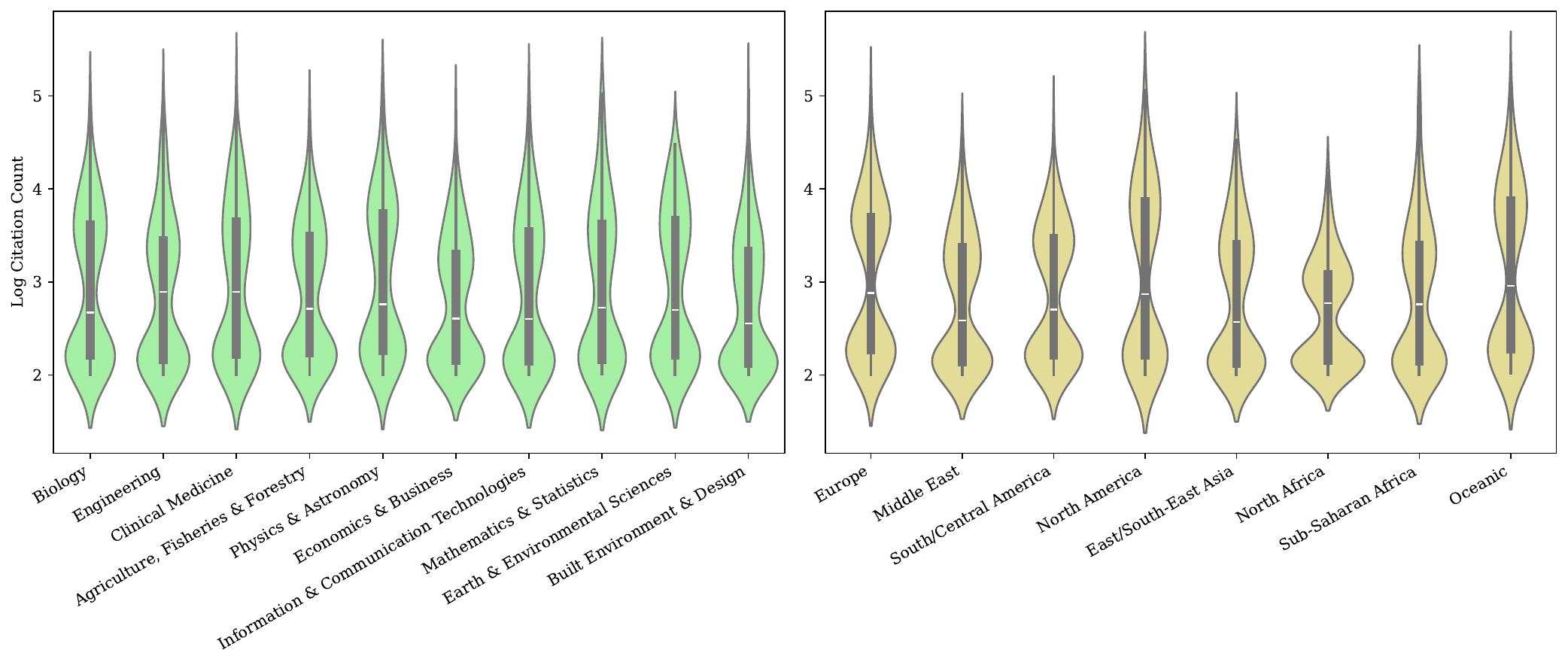}
      \caption{Violin plots of the logarithmic distribution of authors' citation counts across academic disciplines and regions.}
      \label{fig:violinPlots}
  \end{figure*}  

Figure~\ref{fig:violinPlots} {illustrates the variation in authors' citation counts across different fields and regions. Since citation counts follow a power-law distribution, we apply a logarithmic transformation to enhance the comparability of the violin plots. As shown, Physics \& Astronomy and Mathematics \& Statistics have the highest median citation counts among the fields of science, while the regions of Oceanic, Europe, and North America have the top three median citation counts worldwide, consistent with prior studies~\cite{diop2023research}.

\subsubsection{Baseline data collection}

\paragraph{OpenAlex data}

{We consider OpenAlex our primary baseline for coauthorship data, as it is a comprehensive source of bibliographic information, outperforming even well-known scholarly databases like Scopus and Web of Science in terms of inclusivity and coverage} \cite{alperin2024analysis, culbert2025reference}. {We use the OpenAlex API} \cite{priem2022openalex} {to collect all first-degree coauthors of the original 1,600 seed authors. Since OpenAlex profiles are retrieved based on name queries, we verify each match by comparing the affiliation and fields of interest in the OpenAlex profile with those of the original seed author, to ensure we are targeting the correct individual. For 4 of these seed authors, OpenAlex does not return any results, so we exclude them from the analysis, resulting in a final set of 1,596 seed authors.}

\paragraph{Google Scholar data}

{We use Google Scholar as our complementary baseline, as it offers broader citation coverage and contains self-reported coauthorship information, authors manually verify their list of coauthors on the platform} \cite{culbert2025reference, kalhor2022new}. {Using the 1,596 seed authors, we query Google Scholar via the Scholarly API and collect all their first-degree coauthors.}


\subsection{Memorization Metric}\label{subsec:memorizationMetric}
We next describe how memorization has been operationalized in prior work and introduce a metric tailored to our setting. Prior studies have primarily quantified memorization in LLMs through \emph{discoverable extraction}, which measures whether a model can reproduce verbatim segments of its training data when prompted with partial inputs \cite{carlini2021extracting,carlini2022quantifying,nasr2023scalable,hayes2025measuring}. {While effective for detecting exact text memorization, such prefix--suffix formulations are not directly applicable to relational outputs such as unordered lists of entity names.} We therefore adapt the core intuition of discoverable extraction to define a set-based memorization metric suitable for coauthor lists.

\paragraph{\textbf{Definition: Discoverable Name Extraction (DNE)}}

Let \( s \) denote a seed author. Let \( B_s \) be the set of coauthor last names associated with \( s \) in a bibliographic baseline (e.g., Google Scholar or OpenAlex), and let \( G_s \) be the set of coauthor last names generated by an LLM when prompted with \( s \). We define a name-matching function \( m(x,y) \in \{0,1\} \) that returns 1 if the Levenshtein similarity between names \( x \) and \( y \) exceeds a threshold \( \tau \), and 0 otherwise. A baseline coauthor \( b \in B_s \) is considered \emph{discovered} if there exists a generated name \( g \in G_s \) such that \( m(b,g)=1 \). The Discoverable Name Extraction score for author \( s \) is then defined as:

\[
\mathrm{DNE}(s) = \frac{1}{|B_s|} \sum_{b \in B_s} \mathbb{I}\left[ \max_{g \in G_s} m(b,g) = 1 \right],
\]

where \( \mathbb{I}[\cdot] \) is the indicator function. When Google Scholar is used as the baseline, \( |B_s| \) corresponds to the total number of coauthors listed for author \( s \). When OpenAlex is used, we normalize by the minimum of the coauthor counts reported by Google Scholar and OpenAlex to account for coverage differences between databases. To accommodate minor spelling variations and diacritics, particularly for non-English names, we compute name similarity using Levenshtein distance and set the threshold to \( \tau = 0.6 \).\footnote{A threshold of 0.5 approximates random similarity, while higher thresholds were found to be overly restrictive and to underestimate name matches in practice \cite{kalhor2025measuring}.} Based on prior findings on memorization, we expect DNE to be higher for highly cited authors due to their repeated appearance in bibliographic references, and to increase with model size as a function of representational capacity.

\subsection{LLMs}
We examine three LLMs that differ in model size and are likely trained on distinct datasets. Due to the opacity surrounding their training data, assessing the presence of sociopolitical or cultural biases is challenging. While we cannot definitively determine the content to which these models were exposed, our analysis aims to uncover patterns of digital inequality embedded in their outputs.

\textbf{DeepSeek R1:} We include DeepSeek R1~\cite{DeepSeekR1-2025} as an example of an extremely large model, with 671 billion parameters, though only a fraction, 37 billion, are activated per token, optimizing computational efficiency. This model employs multi-stage training and cold-start data augmentation prior to reinforcement learning, achieving reasoning performance comparable to OpenAI-o1. While its base training dataset is unknown, DeepSeek R1 was trained using a unique method called pure reinforcement learning (RL), which does not rely on traditional labeled data.

\textbf{Llama 4 Scout:} We use {Llama 4 Scout}~\cite{meta2025llama4}, which has 17 billion parameters, as an example of a {large-size} LLM. Llama-based models, developed by Meta, have demonstrated advanced language understanding and are trained on large, diverse datasets~\cite{nguyen2023finetuningllama2large,luong-etal-2024-realistic}. While the exact sources of the Llama 4 Scout training dataset are unknown, it is known that Llama models are trained on a mixture of publicly available data, including Common Crawl. This dataset likely contains a substantial amount of scientific literature scraped from open-access journals, institutional repositories, and academic websites, making it plausible that coauthorship and citation patterns from the scientific community are indirectly encoded in the model.

\textbf{Mixtral 8x7B:} Finally, we include Mixtral 8x7B~\cite{Mixtral8x7B-2024}, created by Mistral AI, as an example of a medium-size LLM with 7 billion parameters. This sparse mixture-of-experts model offers strong performance that rivals or exceeds GPT-3.5. Mixtral 8x7B is likely to include scientific literature in its training data, as it is trained on a diverse and extensive web-scale corpus that typically includes open-access academic texts, preprints, and research-related web content.

\subsection{Prompting}
To retrieve coauthorship data from LLMs, we apply zero-shot extraction as described in Section \ref{subsec:memorizationMetric}. {We rely on the default parameter settings (e.g., temperature) provided by each API, as the platforms do not allow for manual adjustment at the time of access.} Each prompt is evaluated independently, with the LLMs generating responses without performing web searches. This setup is essential to our study, as it isolates potential biases inherent to generative AI from those introduced by current search engine outputs. Moreover, since our focus is on the reconstruction of (co-)authors' names, we do not include the titles of publications in the prompt. However, we provide context regarding the seed author's field to reduce ambiguity in the results. The full prompt template is presented in Section \ref{sec:prompt}.



We filter out responses in which the LLM either explicitly indicates that it cannot provide an answer (null responses) or generates {ungrounded} outputs, i.e., {coauthor lists that are not supported by the baseline data.} Examples of responses classified as {ungrounded} and of null responses are provided in Tables~\ref{tab:ungrounded} and~\ref{tab:na}.}




\section{Results}

\subsection{Memorization effect on coauthor lists}

In order to assess the effect of memorization, we first ask: \textit{To what extent does the DNE rate increase as a function of the number of training examples?} We conduct one-sided independent samples \textit{t}-tests on DNE between highly cited and low-cited authors for each of the three LLMs. Our null hypothesis ($H_0$) states that highly cited authors have a higher mean extraction compared to low-cited authors. For all models, our results, at the 0.01 significance level, indicate that highly cited authors have significantly higher extraction than low-cited authors. As shown in Table~\ref{tab:DNEresults}, {DNE scores are generally higher when OpenAlex is used as the baseline. {This aligns with the substantial divergence between the two databases: on average, Google Scholar lists 14.73 coauthors per author, whereas OpenAlex lists 171.64, and the mean Jaccard similarity} \cite{jaccard1901etude} {between their coauthor lists is only 0.08.} This outcome is expected, given that OpenAlex has broader coverage and more complete metadata than Google Scholar. While this difference reflects the comprehensiveness of the underlying reference datasets, the relative gap between highly and low-cited authors remains consistent across both baselines.} These findings suggest that the magnitude of memorization impact in the training sets of LLMs is bounded by a factor of two. That is, as expected, highly cited authors tend to have at least twice as much representation in the GenAI outputs. We describe the implications of this finding for scientometric tools in the Discussion section.

Next, we address the question: \textit{To what extent does the DNE increase as a function of model size?} Our results in Table~\ref{tab:DNEresults} partially support prior findings~\cite{carlini2021extracting} that larger models tend to exhibit higher extraction rates, particularly in the case of DeepSeek R1, which shows the highest DNE scores overall and for highly cited authors. However, this trend is not strictly consistent across all models. For instance, Mixtral 8x7B outperforms Llama 4 Scout in both baselines, suggesting that factors beyond model size---such as training data composition, architectural choices, or alignment strategies---also influence memorization behavior. {Importantly, the effect of model size is asymmetric across citation strata. While larger models yield higher DNE scores for highly cited authors, we observe uniformly low DNE values and no clear scaling trend for low-cited authors. This pattern suggests that increasing model capacity primarily amplifies memorization when sufficient exposure exists in the training data, whereas authors with sparse representation and shorter coauthor lists do not benefit substantially from larger model sizes.} Table~\ref{tab:DNE_eps_combined} reports results for higher DNE thresholds, which remain consistent with our main findings.


\begin{table*}[!ht]
\centering
\caption{DNE results across two baselines: OpenAlex and Google Scholar. For each LLM, we report mean DNE scores for high- and low-cited authors, t-test results, and overall mean ± SD. p-values $<$ 0.05 are indicated with *, $<$ 0.01 with **, and $<$ 0.001 with ***.}
\resizebox{\textwidth}{!}{
\begin{tabular}{l|c|ccccc|ccccc}
\toprule
\multirow{2}{*}{\textbf{LLM}} & \multirow{2}{*}{\textbf{Params}} 
& \multicolumn{5}{c|}{\textbf{OpenAlex as Baseline}} 
& \multicolumn{5}{c}{\textbf{Google Scholar as Baseline}} \\
\cmidrule(lr){3-7} \cmidrule(lr){8-12}
& & DNE$_\text{High}$ & DNE$_\text{Low}$ & T-Stat & P-Val & Overall (±SD)
  & DNE$_\text{High}$ & DNE$_\text{Low}$ & T-Stat & P-Val & Overall (±SD) \\
\midrule
DeepSeek R1      & 671B & 0.70 & 0.35 & 21.04 & *** & 0.54 (0.35)
                        & 0.21 & 0.09 & 14.02 & *** & 0.15 (0.17) \\
Llama 4 Scout    & 17B  & 0.49 & 0.32 & 5.51  & *** & 0.44 (0.34)
                        & 0.12 & 0.06 & 4.89  & *** & 0.10 (0.14) \\
Mixtral 8x7B     & 7B   & 0.63 & 0.35 & 16.66  & *** & 0.49 (0.36)
                        & 0.15 & 0.08 & 8.57  & *** & 0.12 (0.16) \\
\bottomrule
\end{tabular}
}
\label{tab:DNEresults}
\end{table*}

\subsection{Impact of memorization across academic disciplines}

In this section, we assess the effect of memorization across different academic disciplines. According to our hypothesis, memorization is evenly distributed among various fields. In other words, we hypothesize that LLMs return more accurate results for highly cited researchers compared to low-cited researchers, regardless of their field of study. {To test this hypothesis, we perform a t-test within each academic discipline for each LLM–baseline pair.} Figure~\ref{fig:radarPlotsDiscipline} displays the average DNE across ten academic disciplines {for each combination of LLM and baseline,} disaggregated by citation level. The results show that, for Llama 4 Scout, {several fields exhibit significant effects of memorization when compared against both OpenAlex and Google Scholar, though the specific disciplines affected vary by baseline. In particular, Earth \& Environmental Sciences emerges as a consistent area of concern, showing significant memorization effects in both comparisons. When compared to Google Scholar, additional fields such as Built Environment \& Design and Economics \& Business show strong memorization effects, whereas in the OpenAlex comparison, the fields of Engineering, Clinical Medicine, and Biology demonstrate significant impacts. Information and Communication Technologies shows a trend toward memorization in the OpenAlex comparison, while Engineering and Mathematics \& Statistics trend similarly in the Google Scholar comparison.} However, the {coauthor lists} of low-cited researchers in other fields are better represented and exhibit less bias compared to the aforementioned disciplines. 

{For DeepSeek R1 and Mixtral 8x7B, the pattern differs from that of Llama 4 Scout, with memorization effects observed across a broader range of disciplines. When compared with OpenAlex, both models exhibit significant memorization effects across all fields, indicating a widespread tendency to reproduce or overfit coauthorship patterns seen in the training data.} In contrast, when compared with Google Scholar, DeepSeek R1 shows significant effects in all fields except Clinical Medicine, while Mixtral 8x7B shows memorization in most fields, with Clinical Medicine, Agriculture, Fisheries \& Forestry, Biology, and Physics \& Astronomy showing only a trend toward significance. Despite these widespread effects, DeepSeek R1 consistently outperforms the other LLMs in generating {coauthor lists} for highly cited authors, producing fewer incorrect coauthors across {both baselines} and research domains.

\begin{figure*}[!ht]
    \centering
   \includegraphics[width=\linewidth]{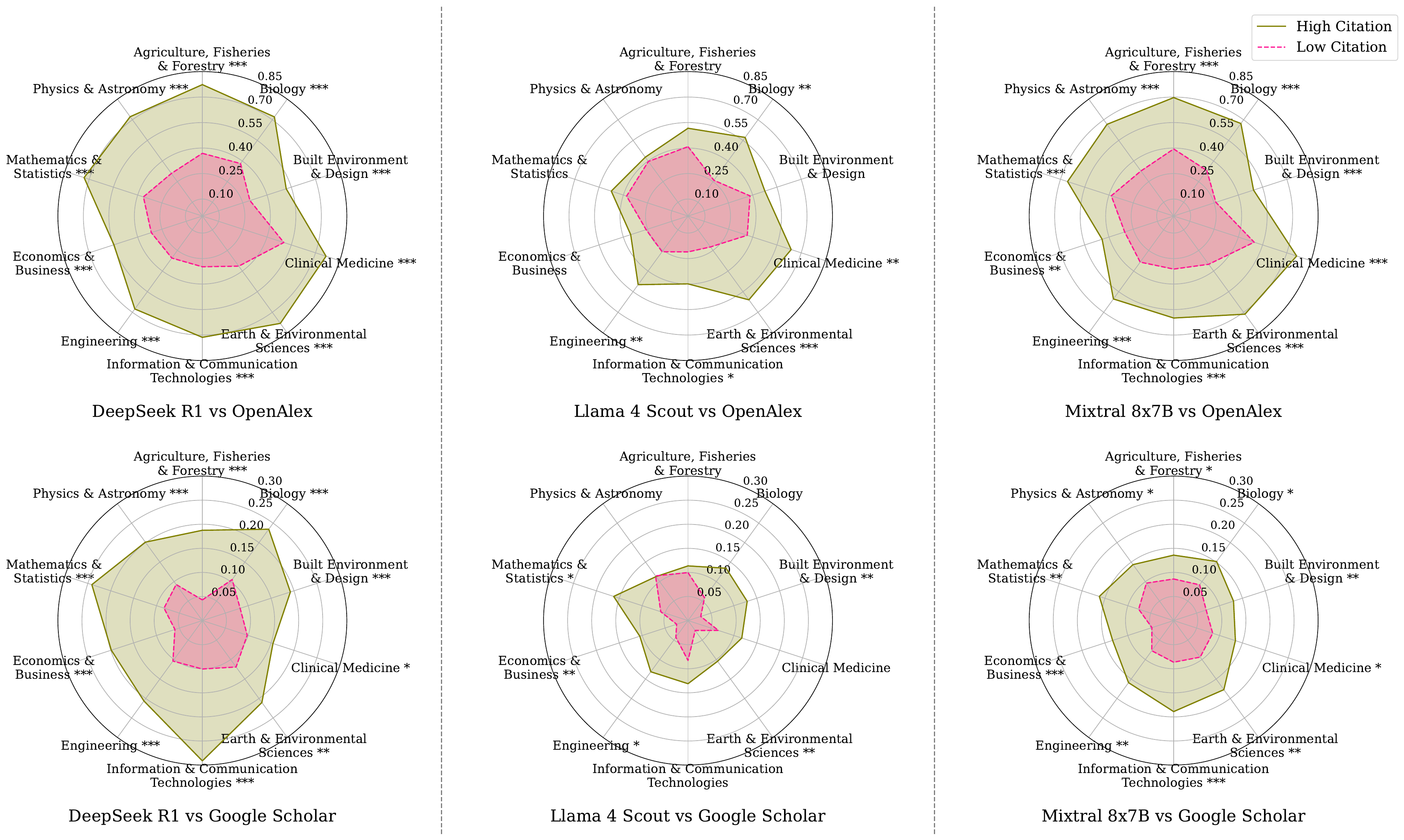}
    \caption{
        Mean DNE for highly cited authors (olive) and low-cited authors (pink) across different academic disciplines, shown for each LLM–baseline pair. To improve visual clarity, different scales are used for the two baselines. p-values $<$ 0.05 are indicated with *, p-values $<$ 0.01 with **, and p-values $<$ 0.001 with ***.
    }
    \label{fig:radarPlotsDiscipline}
\end{figure*}

{Additionally, we compare the memorization effect across academic disciplines.} Figure~\ref{fig:groupedBarDiscipline} illustrates the mean DNE across academic disciplines for each LLM, {presented separately by baseline.} Based on this figure, DeepSeek R1 consistently achieves higher DNE values across nearly all fields, {with the exception of Built Environment \& Design when using OpenAlex as the baseline}, which aligns with its larger number of parameters. The difference in mean DNE between Llama 4 Scout and Mixtral 8x7B is smaller, reflecting the narrower gap in their parameter sizes. The lower DNE values for Llama 4 Scout may be attributed to the fewer coauthors it generates compared to the other models. {Furthermore, when comparing mean DNE scores across baselines, we observe, consistent with previous sections, substantially higher values for OpenAlex than for Google Scholar, indicating a stronger memorization signal when OpenAlex is used as the reference set.}

\begin{figure*}[!ht]
    \centering
   \includegraphics[width=\linewidth]{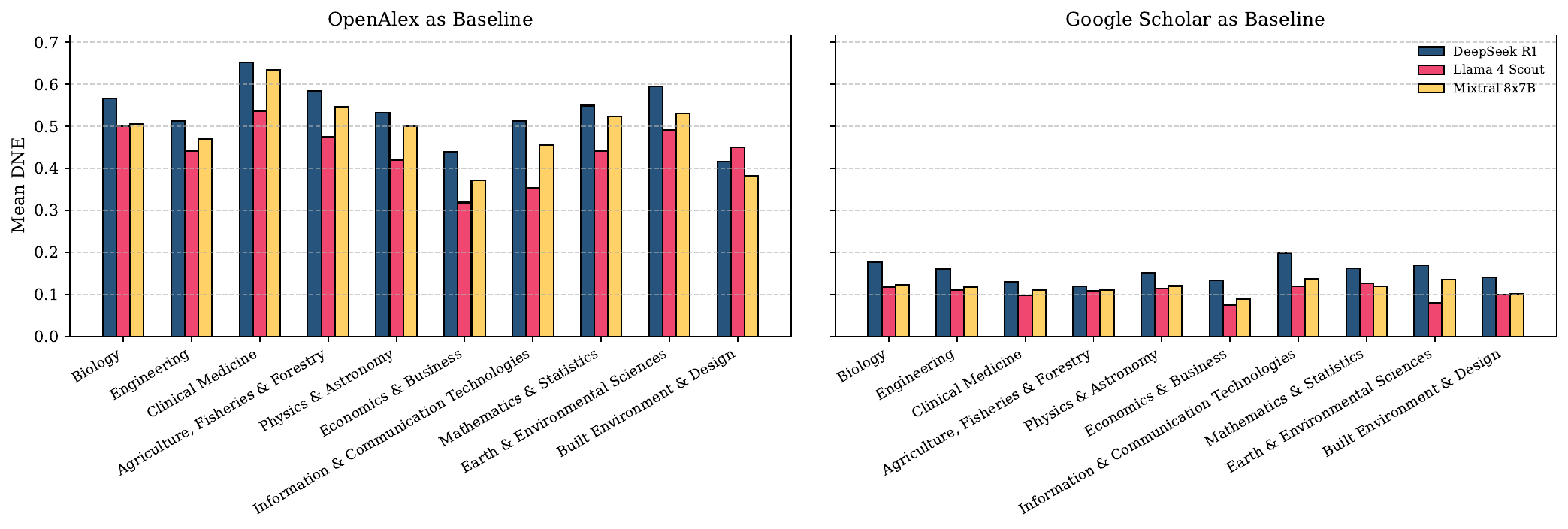}
    \caption{
        {Grouped bar plots showing the mean DNE across academic disciplines for each LLM, separated by baseline.}
    }
    \label{fig:groupedBarDiscipline}
\end{figure*}

\subsection{Impact of memorization across regions}

We next explore how memorization varies across regions, testing the hypothesis that it is evenly distributed worldwide. Specifically, we expect higher DNE for highly cited authors compared to their low-cited counterparts, regardless of geographic location. Figure~\ref{fig:radarPlotsRegion} shows average DNE by region {for each combination of LLM and baseline,} disaggregated by citation level. For Llama 4 Scout, East/South-East Asia stands out with the strongest memorization effects. Significant memorization appears in East/South-East Asia and Sub-Saharan Africa {across both baselines.} Additionally, we observe trends toward significance in the Middle East, South/Central America, and Oceanic, {though notably, the Oceanic trend appears only in the Google Scholar comparison, not in the OpenAlex-based analysis. This indicates some regional sensitivity in the model's memorization patterns depending on the baseline used.}

In contrast, DeepSeek R1 exhibits its strongest memorization effects in North America, with all regions showing significant effects {in the OpenAlex comparison. However, in the Google Scholar comparison,} Sub-Saharan Africa shows only a trend toward significance. This suggests that the scholarly output of low-cited authors from Sub-Saharan Africa is relatively well represented in DeepSeek R1's training data, resulting in reduced representational bias, {at least when measured against Google Scholar.} For Mixtral 8x7B, the most pronounced memorization is observed in South/Central America. All regions show significant effects {in the OpenAlex comparison, while in the Google Scholar comparison,} North Africa exhibits only a trend toward significance. This implies that the work of low-cited authors from North Africa is comparatively well captured in Mixtral 8x7B's training data, again reflecting lower representational bias in that region {under the Google Scholar baseline.}

\begin{figure*}[!ht]
    \centering
   \includegraphics[width=\linewidth]{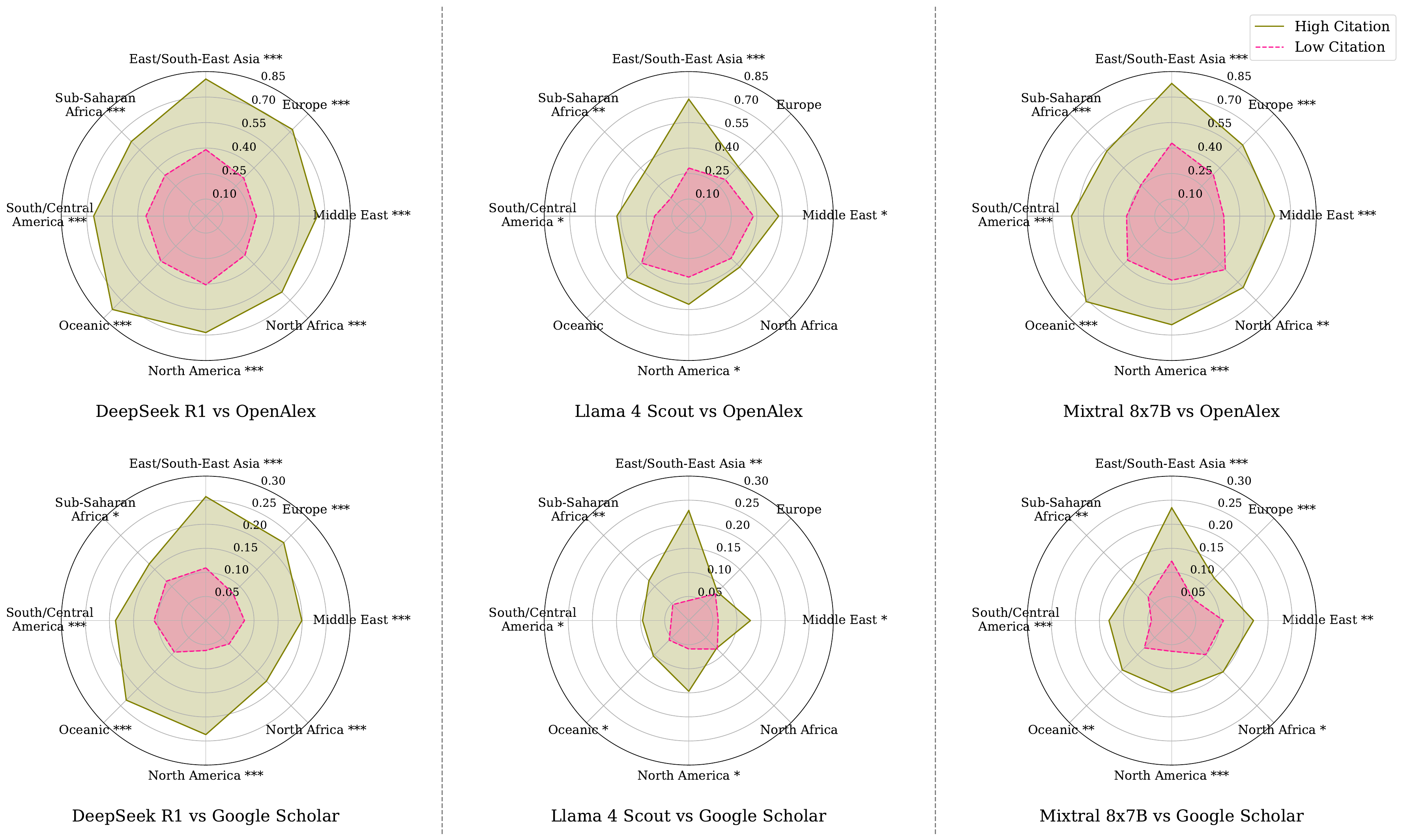}
    \caption{
        Mean DNE for highly cited authors (olive) and low-cited authors (pink) across different regions, shown for each LLM–baseline pair. To improve visual clarity, different scales are used for the two baselines. p-values $<$ 0.05 are indicated with *, p-values $<$ 0.01 with **, and p-values $<$ 0.001 with ***.
    }
    \label{fig:radarPlotsRegion}
\end{figure*}

{We also examine how the memorization effect varies across geographic regions.} Figure~\ref{fig:groupedBarRegion} shows the mean DNE across eight regions for each LLM, {split by baseline.} Similar to the previous section, DeepSeek R1 achieves the highest values for most regions. An exception occurs in East/Southeast Asia {under the Google Scholar baseline,} where Llama 4 Scout outperforms. Additionally, in North Africa, Mixtral 8x7B achieves the highest mean DNE. These exceptions aside, the overall trend reinforces the impact of model size (i.e., number of parameters) on achieving higher mean DNE. For Llama 4 Scout and Mixtral 8x7B, the mean DNEs are close, with Mixtral 8x7B being slightly higher in a few cases. This is expected, as Llama 4 Scout tends to return fewer coauthors than requested in the prompts, despite having more parameters. {Moreover, in line with the findings from RQ1 and RQ2, using OpenAlex as the baseline yields much higher mean DNE values than Google Scholar, reflecting the stronger memorization signal captured by the broader OpenAlex dataset.}

\begin{figure*}[!ht]
    \centering
   \includegraphics[width=\linewidth]{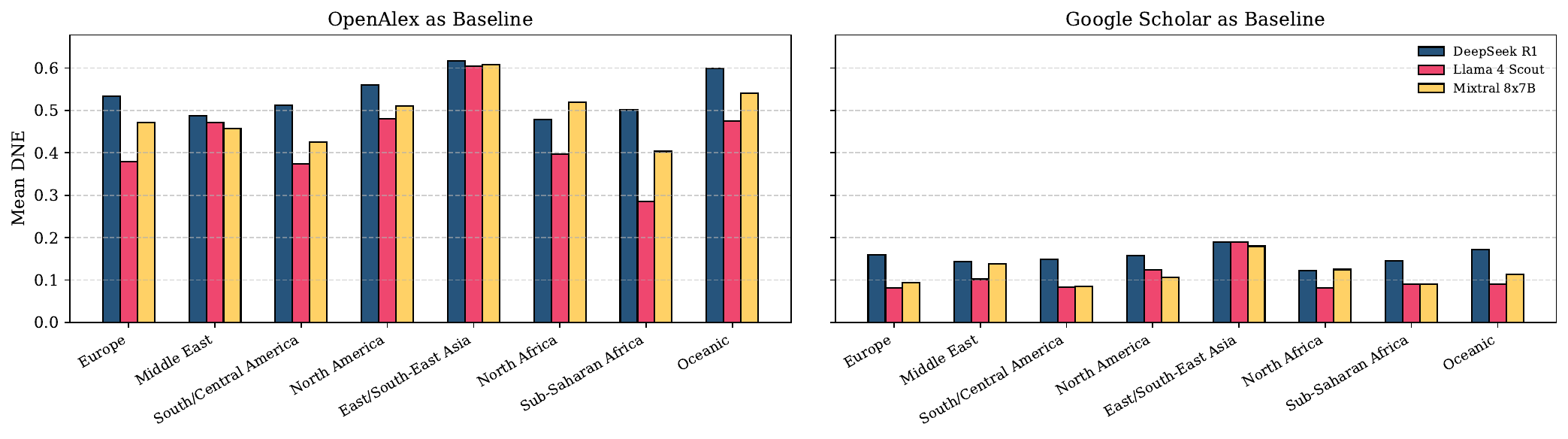}
    \caption{
        {Grouped bar plots showing the mean DNE across regions for each LLM, separated by baseline.}
    }
    \label{fig:groupedBarRegion}
\end{figure*}

\section{Discussion}

Our study examines LLMs' memorization capabilities in representing {coauthor lists} and investigates biases across different academic disciplines and regions. The findings highlight LLMs' strengths and limitations in scientometric applications, offering valuable insights into their responsible use.

\subsection{Memorization effects}

While LLMs do not explicitly ``memorize'' {coauthor lists}, they implicitly capture relationships between authors, papers, and topics based on the co-occurrence of names, affiliations, and keywords during training. Our results confirm that the learned associations, shaped by statistical patterns in the training data, are overfitted toward highly cited, frequently mentioned researchers. {This asymmetry also helps explain why increasing model size improves DNE primarily for highly cited authors: when authors appear rarely in training data and tend to have short coauthor lists, as is typical for low-cited researchers, additional model capacity does not substantially enhance memorization.} This bias leads to less accurate reconstruction of {coauthor lists} for less-cited authors or those from underrepresented disciplines or regions. When LLMs are used to generate or summarize such {relational information}, these inaccuracies can reduce the visibility of these researchers in {scholarly discovery contexts.}

Tackling biases in scientometric systems that integrate LLMs is a particularly complex challenge. Unlike traditional information retrieval (IR) systems, which can adapt to new citations or update their link structures, LLMs require substantial retraining to unlearn deeply ingrained associations. These biases are often embedded at the level of token representations. Furthermore, the initial advantage enjoyed by highly cited authors is not solely a consequence of their citation counts; it is often amplified by broader systemic factors. Their work is more likely to be open access (partly due to their ability to cover publication fees), presented at prominent conferences, and better indexed in traditional search engines. This creates a ``rich get richer'' dynamic, where a small group of highly visible researchers not only dominate citation networks but also enjoy disproportionate representation in the next generation of AI-driven scientific systems, further entrenching existing inequalities.

\subsection{Disparities across academic disciplines and regions}

While our global analysis (RQ1) indicates a consistent memorization bias favoring highly cited authors, this pattern does not hold uniformly across academic disciplines and geographical regions. A more granular analysis reveals that disparities in LLM memorization are shaped by both the field of study and regional representation in training data. Across disciplines, we observe that certain fields, such as Engineering, Built Environment \& Design, and Economics \& Business, exhibit particularly strong memorization effects, with {coauthor lists associated with} highly cited authors reproduced far more frequently than those of low-cited peers. In contrast, Clinical Medicine shows a much narrower gap, suggesting that its complex and heterogeneous {collaboration patterns} and publication practices temper the model's ability recall high-visibility {coauthor information}. These discipline-specific differences imply that both the typical {scale of collaboration} and the digital footprint of a field critically influence how LLMs internalize scholarly relationships.

Geographically, memorization effects also vary considerably. DeepSeek R1 consistently favors highly cited authors in regions with extensive, digitized literature, such as North America and Europe, whereas in under-represented areas like Sub-Saharan Africa the gap between high- and low-cited cohorts narrows to near parity. Mixtral 8×7B similarly exhibits reduced representational bias in North Africa, and Llama 4 Scout, despite lower overall extraction, yields almost no difference between citation groups in Europe and {North Africa}. These regional nuances likely reflect uneven inclusion of non-Western and low-resource scholarship in publicly available training corpora. Together, our findings highlight persistent equity challenges in LLM-driven scholarly analyses and underscore the need for more inclusive data collection and modeling strategies.

\subsection{Implications for scientometric applications}

{We do not argue that LLMs should replace curated bibliographic databases or paper-level metadata for formal scientometric analysis. Rather, our focus is on the growing role of LLMs as intermediate reasoning layers within scholarly discovery tools, recommendation systems, and AI-assisted research workflows. In these settings, LLMs often generate relational outputs---such as related authors, collaboration patterns, or expertise signals---based on their internal representations rather than direct database queries. Evaluating how accurately and equitably these representations encode coauthorship relations is therefore relevant to both scientometric integrity and fair scholarly information access.}

The findings from this study represent an important step toward {understanding the implications of LLM-mediated outputs for scientometric analysis, particularly when coauthor lists are implicitly reconstructed in scholarly discovery and recommendation systems.} By uncovering disparities and biases inherent in LLM-generated outputs, we identify critical opportunities to refine these systems for more equitable and accurate academic representations. Specifically, we demonstrate that LLMs disproportionately favor well-known authors and dominant regions, highlighting the urgent need for interventions to address these imbalances. These findings suggest that while LLMs can automate aspects of {generating relational scholarly information}, their outputs must be critically evaluated to avoid perpetuating or amplifying existing inequities.

The ability of LLMs to process vast datasets and infer relationships beyond explicit coauthorship underscores their utility for exploratory analyses. However, their interpretative biases necessitate careful calibration. Integrating LLMs into scientometric workflows has the potential to enhance the detection of emerging trends, identify underrepresented collaborations, and enrich bibliometric studies, provided these tools are supported by robust validation methods and a commitment to mitigating inherent biases.

In light of these findings, our work underscores the need for an interdisciplinary approach to improving LLMs for academic applications. Collaboration among AI practitioners, domain experts, and policymakers is essential to ensure that scholarly tools reflect the diverse and interconnected nature of the global academic community. 

\subsection{Limitations and future directions}
This study is subject to several limitations that suggest directions for future research. {First, there is a temporal mismatch between continuously updated bibliographic databases and the fixed training cutoffs of LLMs. While resources such as Google Scholar and OpenAlex regularly incorporate newly published papers and evolving coauthorships, LLMs only encode information available up to the end of their training period. Consequently, low DNE values may reflect genuinely missing or outdated information rather than a failure of memorization or representational bias. This effect is likely to be more pronounced for early-career researchers, rapidly evolving fields, or authors whose most influential collaborations emerged after the model's training cutoff.}

{Second, this study does not explicitly resolve author name disambiguation, which remains a well-known and challenging problem in bibliographic analysis. While we rely on author-curated Google Scholar profiles as a baseline to reduce ambiguity, the LLM prompts intentionally do not include publication titles or unique identifiers, in order to preserve a zero-shot setting and avoid injecting database-specific metadata into the generation process. As a result, authors with common names or overlapping research areas may be conflated in LLM-generated outputs, potentially leading to under- or over-estimation of DNE in some cases. This limitation is particularly relevant for fields with dense collaboration networks or high name ambiguity. Addressing fine-grained author disambiguation---e.g., through controlled identifiers or publication-level context—remains an important direction for future work.}

Third, the {baseline} {coauthor lists} used for comparison may themselves reflect biases inherent in citation-based metrics, potentially affecting the interpretation of observed discrepancies. Exploring alternative benchmarks, such as ethnographic insights or manually curated datasets, could provide a more nuanced evaluation. Additionally, the analysis primarily focuses on high-level patterns, leaving the influence of other important factors, such as the gender of authors and coauthors, underexplored. 

Fourth, while this study primarily examines the representational biases of LLMs, future work could develop benchmark comparison datasets to systematically measure and mitigate these biases in downstream tools. For instance, one potential approach might involve reverting to traditional search methods in cases where algorithmic bias is likely. Such advancements could enhance the reliability and fairness of scientometric tools built on LLMs.

\section{Conclusion}

In this paper, we analyze the influence of LLM memorization on {coauthor lists} using three well-known models: DeepSeek R1, Llama 4 Scout, and Mixtral 8x7B. By employing Discoverable {Name} Extraction as our memorization metric, we find that memorization effects are significantly stronger for highly cited authors compared to their lower-cited counterparts, as evidenced by the more accurate and frequent representation of these researchers in LLM-generated {coauthor lists}. Additionally, we observe that models with a greater number of parameters tend to exhibit a higher degree of memorization in these lists. However, there are notable exceptions to this overall pattern when examined across academic disciplines and regions. At the disciplinary level, Clinical Medicine shows no significant difference in DNE scores between highly and lower-cited authors. At the regional level, areas such as Sub-Saharan Africa and North Africa demonstrate near-equitable representation of highly and lower-cited scholars. These insights highlight the importance of developing more inclusive and equitable approaches to leveraging LLMs in {scholarly discovery and representation}. {Importantly, these findings should not be interpreted as advocating the use of LLMs as substitutes for curated bibliographic databases, but rather as highlighting how LLM-mediated scholarly interfaces can implicitly shape the visibility and representation of academic relationships.}




\backmatter

\section*{Abbreviations}
  LLMs, Large Language Models; DNE, Discoverable {Name} Extraction; RL, Reinforcement Learning; IR, Information Retrieval.

\section*{Declarations}

\subsection*{Availability of data and materials}

The data and analysis code that support the findings of this study are available at \url{https://github.com/kalhorghazal/LLMs-Co-Authorship-Memorization}.

\subsection*{Competing interests}

The authors declare that they have no competing interests.

\subsection*{Funding}

No funding was received for conducting this study.

\subsection*{Authors' contributions}

GK: Data curation, Formal analysis, Methodology, Investigation, Validation, Visualization, Writing- Original draft. AM: Conceptualization, Project administration, Supervision, Writing- Reviewing and Editing. All authors read and approved the final manuscript.




\begin{appendices}

\section{Prompt template}\label{sec:prompt}

\begin{graybox}
\textbf{Prompt:} \textit{Can you list the top coauthors of {\color{blue}[name from Google Scholar profile]}, who works in the field of {\color{blue}[field from Google Scholar profile]}? Please provide the full names (first and last) of up to {\color{blue}[number of coauthors from Google Scholar]} coauthors. Separate each coauthor's full name from the next using a forward slash ('/'), without adding extra whitespace.
}
\end{graybox}

{All prompt parameters (author name, field, and list length) are populated externally from Google Scholar and are not disclosed to the LLM as database-specific information.}

\section{DNE results for different thresholds}
\begin{table*}
\caption{DNE results across different thresholds ($\epsilon = 0.7, 0.8, 0.9$) and two baselines: \textbf{OpenAlex} and \textbf{Google Scholar}. For each LLM, we report DNE scores for high- and low-cited authors, t-test results, and overall mean ± SD. p-values $<$ 0.05 are indicated with *, $<$ 0.01 with **, and $<$ 0.001 with ***.}
\centering
\resizebox{\textwidth}{!}{
\begin{tabular}{l|ccccc|ccccc}
\toprule
\multirow{2}{*}{\textbf{LLM}} 
& \multicolumn{5}{c|}{\textbf{OpenAlex as Baseline}} 
& \multicolumn{5}{c}{\textbf{Google Scholar as Baseline}} \\
\cmidrule(lr){2-6} \cmidrule(lr){7-11}
& DNE$_\text{High}$ & DNE$_\text{Low}$ & T-Stat & P-Val & Overall (±SD)
& DNE$_\text{High}$ & DNE$_\text{Low}$ & T-Stat & P-Val & Overall (±SD) \\
\midrule
\multicolumn{11}{c}{\textbf{$\epsilon = 0.7$}} \\
\midrule
DeepSeek R1      & 0.47 & 0.18 & 19.13 & *** & 0.33 (0.32)
                 & 0.14 & 0.04 & 13.55 & *** & 0.09 (0.14) \\
Llama 4 Scout    & 0.29 & 0.12 & 6.26 & *** & 0.24 (0.28)
                 & 0.06 & 0.02 & 4.01 & *** & 0.05 (0.10) \\
Mixtral 8x7B     & 0.37 & 0.17 & 12.64 & *** & 0.27 (0.31)
                 & 0.07 & 0.03 & 7.25 & *** & 0.05 (0.11) \\
\midrule
\multicolumn{11}{c}{\textbf{$\epsilon = 0.8$}} \\
\midrule
DeepSeek R1      & 0.38 & 0.12 & 18.23 & *** & 0.26 (0.29)
                 & 0.12 & 0.03 & 13.12 & *** & 0.08 (0.13) \\
Llama 4 Scout    & 0.21 & 0.08 & 5.18 & *** & 0.17 (0.25)
                 & 0.06 & 0.02 & 3.67 & *** & 0.03 (0.09) \\
Mixtral 8x7B     & 0.26 & 0.12 & 9.82 & *** & 0.19 (0.28)
                 & 0.05 & 0.02 & 6.65 & *** & 0.03 (0.09) \\
\midrule
\multicolumn{11}{c}{\textbf{$\epsilon = 0.9$}} \\
\midrule
DeepSeek R1      & 0.33 & 0.10 & 16.80 & *** & 0.23 (0.28)
                 & 0.11 & 0.03 & 12.75 & *** & 0.07 (0.12) \\
Llama 4 Scout    & 0.16 & 0.06 & 4.52 & *** & 0.13 (0.23)
                 & 0.04 & 0.01 & 3.18 & *** & 0.03 (0.08) \\
Mixtral 8x7B     & 0.20 & 0.09 & 8.35 & *** & 0.15 (0.26)
                 & 0.03 & 0.01 & 6.30 & *** & 0.02 (0.07) \\
\bottomrule
\end{tabular}
}
\label{tab:DNE_eps_combined}
\end{table*}

\section{Field and subfield values}
\begin{table*}
\centering
\footnotesize
\caption{Fields and their corresponding subfields in our dataset.}\label{tab:field}
\renewcommand{\arraystretch}{1.2} 
\begin{tabular}{p{0.15\linewidth}p{0.8\linewidth}}
\toprule
\textbf{Field} & \textbf{Subfields}\\
\midrule
Agriculture, Fisheries \& Forestry & 
Agronomy \& Agriculture/Dairy \& Animal Science/Fisheries/Food Science/ Forestry/Horticulture/Veterinary Sciences\\
\midrule
Built Environment \& Design & 
Architecture/Building \& Construction/Design Practice \& Management/Urban \& Regional Planning\\
\midrule
Engineering & 
Aerospace \& Aeronautics/Automobile Design \& Engineering/Biomedical Engineering/Chemical Engineering/Civil Engineering/Electrical \& Electronic Engineering/Environmental Engineering/Geological \& Geomatics Engineering/Industrial Engineering \& Automation/Mechanical Engineering \& Transports/Mining \& Metallurgy/Operations Research\\
\midrule
Information \& Communication Technologies & 
Computation Theory \& Mathematics/Computer Hardware \& Architecture/ Distributed Computing/Image Processing/Information Systems/Medical Informatics/Networking \& Telecommunications/Software Engineering\\
\midrule
Economics \& Business & 
Accounting/Agricultural Economics \& Policy/Business \& Management/ Development Studies/ Econometrics/Economic Theory/Economics/Finance/Industrial Relations/Logistics \& Transportation/Sport, Leisure \& Tourism\\
\midrule
Clinical Medicine & 
Allergy/Anesthesiology/Arthritis \& Rheumatology/Cardiovascular System \& Hematology/ Complementary \& Alternative Medicine/Dentistry/Dermatology \& Venereal Diseases/Emergency \& Critical Care Medicine/Endocrinology \& Metabolism/Environmental \& Occupational Health/Gastroenterology \& Hepatology/General \& Internal Medicine/General Clinical Medicine/Geriatrics/Legal \& Forensic Medicine/Neurology \& Neurosurgery/Obstetrics \& Reproductive Medicine/Ophthalmology \& Optometry/Orthopedics/Otorhinolaryngology/ Pathology/Pediatrics/Pharmacology \& Pharmacy/Respiratory System/Sport Sciences/Surgery/Tropical Medicine/Urology \& Nephrology\\
\midrule
Biology & 
Entomology/Evolutionary Biology/Marine Biology \& Hydrobiology/ Ornithology/Plant Biology \& Botany/Zoology\\
\midrule
Earth \& Environmental Sciences & 
Environmental Sciences/Geochemistry \& Geophysics/Geology/Meteorology \& Atmospheric Sciences/Oceanography/Paleontology\\
\midrule
Mathematics \& Statistics & 
Applied Mathematics/General Mathematics/Numerical \& Computational Mathematics/Statistics \& Probability\\
\midrule
Physics \& Astronomy & 
Acoustics/Applied Physics/Astronomy \& Astrophysics/Chemical Physics/Fluids \& Plasmas/General Physics/Mathematical Physics/Optics\\
\bottomrule
\end{tabular}
\end{table*}

\section{Null response examples}
\renewcommand{\arraystretch}{1.2} 
{\footnotesize
\begin{longtable}{p{0.15\linewidth} p{0.8\linewidth}}
\caption{Anonymized examples of {ungrounded} responses generated by LLMs.}\label{tab:ungrounded} \\
\toprule
\textbf{LLM} & \textbf{{Ungrounded} Response}\\
\endfirsthead
\toprule
\textbf{LLM} & \textbf{{Ungrounded} Response (continued)}\\
\endhead
\midrule
DeepSeek R1 & [Author’s Name]'s top co-authors in [Author’s Subfield] may include researchers such as (hypothetical examples): Maria Fernandez/Rajesh Kumar/Amina Khaled/Carlos Mendez/Sofia Ivanova/Hiroshi Tanaka/Fatima Al-Mansoori/Samuel Ogunleye/Elena Petrova/Ahmed Hassan/Li Chen/Grace O'Connor/Paolo Russo. Note: These names are illustrative; actual collaborators would require analysis of publication databases.\\
\midrule
DeepSeek R1 & Due to the lack of real-time access to academic databases or publications, I cannot provide verified co-authors for [Author’s Name] in [Author’s Subfield]. For accurate results, consider consulting platforms like Google Scholar, ResearchGate, or Scopus. If you'd like a hypothetical example (not factual), here's a placeholder format:
\par
Antonio Perez-Lopez/Maria Hernandez-Garcia/Carlos Ruiz/Laura Gomez-Fernandez/Diego Silva-Mendez\\
\midrule
DeepSeek R1 & I cannot provide real-time or updated information on specific individuals' co-authors. To find the most accurate and current list of [Author’s Name]'s co-authors, I recommend checking academic databases like PubMed, Google Scholar, or ResearchGate. If you need hypothetical examples for formatting purposes, here’s a generic illustration: John D. Smith/Jane M. Doe. Replace these with verified names from relevant publications.\\
\midrule
DeepSeek R1 & [Author’s Name]'s frequent collaborators in [Author’s Subfield] are not explicitly documented in publicly available sources as of 2023. To ensure accuracy, please consult academic databases (e.g., Scopus, Google Scholar) or institutional profiles for verified co-authorship details. If speculative examples are needed for format purposes, consider placeholder names like John Doe/Jane Smith, but these are not factual.\\
\midrule
DeepSeek R1 & As of my knowledge cutoff in July 2024, there is no widely recognized public record of an economist named [Author’s Name] in prominent academic or research databases. If this is a hypothetical or emerging researcher, specific co-author details cannot be reliably provided. For accurate results, please verify the name’s spelling or context. If hypothetical, here’s an example format: 
\par
Emily Carter/James Rivera/Sophia Lee/Michael Chen\\
\midrule
Llama 4 Scout & To provide an accurate list of top co-authors for [Author’s Name] in the field of [Author’s Subfield], I would need access to specific data or publications related to his work. However, based on general information and assuming a reliance on common academic databases or publication records, I can offer a speculative response. 
\par
Given the constraints and without direct access to current or specific publication data for [Author’s Name], I'll proceed with a hypothetical example. Let's assume [Author’s Name] has collaborated with several individuals in the field of [Author’s Subfield]. Here's a speculative list:
\par
Rajat Pal / Rajesh Gupta / Sumeet Singh
\par
Please note, this is a fictional example. For an accurate and up-to-date list, one would ideally refer to academic databases like Google Scholar, ResearchGate, Academia.edu, or IEEE Xplore, where co-authorship information is regularly updated. If [Author’s Name] has a profile on these platforms, you could find the actual top co-authors by examining his publication history.\\
\midrule
Llama 4 Scout & I don't have access to real-time data or specific information about [Author’s Name]'s co-authors. However, I can suggest some possible sources where you may be able to find this information, such as academic databases like Google Scholar, ResearchGate, or the Scopus database. 
\par
If I had to provide an example answer based on a hypothetical search, it could look something like this:
\par
Darrell J. Wastgård /  Ana Slatnar /  Urszula Gawlik-Dowaniak /  Jernej Pavček /  Luka  Vidovič 
\par
Please verify this information through a reliable source as it may not be accurate or up-to-date.\\
\midrule
Llama 4 Scout & I don't have access to real-time data or specific information about individuals' publications and co-authors. However, I can suggest some possible ways to find the co-authors of [Author’s Name] in the field of [Author’s Subfield]:
\par
1. Search online academic databases such as Google Scholar, ResearchGate, or Academia.edu for [Author’s Name]'s publications and co-authors.
\par
2. Look up [Author’s Name]'s profile on professional networking sites or academic platforms.
\par
That being said, I can provide you with a general response based on common co-authors in the field of [Author’s Subfield]. Here are 18 possible co-authors separated by a forward slash:
\par
Jianlin Cao / Wei Chen / Jianqiang Zhang / Haowei Lai / Qiang Lin / Xiao Wang / Hua Zou / Hongyan Wang / Yong Zhang / Shaoping Wang / Lei Zhang / Jin Shi / Zhenxing Wang / Guofu Zhang / Xiaoming Li / Weining Ni / Yuan Cao / Jianguo Wang 
\par
Please note that these names are randomly generated and may not be actual co-authors of [Author’s Name]. For accurate information, I recommend searching online academic databases or [Author’s Name]'s professional profile.\\
\midrule
Llama 4 Scout & I can provide you with the list of top co-authors. However, I need to access the data on [Author’s Name]'s publications. Assuming I have that data, here is the list:
\par
[Author’s Name] does not have a large number of co-authors that are widely documented. However, based on available data: 
\par
Abdul Haadi /Mohammed Ahmed / Saud Al-jaddid / Hanaa A. Eldin /Mohamed R. Mahmoud / Ali Mohamed /Nabil H. Elhassan /Taghreed M. Hassan /S. M. Elseify /E. A. Elsayed /Sultan J. Al-badrani /Manal M. Saber /Y. W. Eltahir /A. S. Rabie /M. A. W. /A. Eldesoukey /Emad A. /Khalil /H. M. /El-Sayed /A. A. /Mohamed /S. A. /Ahmed /M. S. /El-Sayed /Mohamed S. /A. M. 
\par
Please note that I had to simulate that data as real access to Amal Soliman Hassan's data wasn't available. The correct list would come from an actual publication or author database. Always confirm through reliable sources.\\
\midrule
Llama 4 Scout & I don't have access to real-time information or specific data about individuals' publications or collaborations. However, I can suggest some possible sources where you may be able to find the co-authors of [Author’s Name]:
\par
1. Google Scholar (scholar.google.com)\par
2. ResearchGate (researchgate.net)\par
3. Academia.edu (academia.edu)\par
4. DBLP (dblp.uni-trier.de)\par
\par
If you search for [Author’s Name] on these platforms, you may be able to find his publications and co-authors. 
\par
Since I don't have the information, I'll provide a generic response. If you provide me with the actual data, I can help you format it as requested: 
\par
John Smith / Jane Doe / Bob Johnson / Maria Rodriguez / David Lee / Emily Chen / Kevin White 
\par
Please look up the information from the suggested sources.\\
\midrule
Mixtral 8x7B & I'm unable to browse the web, so I can't provide real-time information about specific individuals. However, I can provide you with a general sense of how to find this information. To find [Author’s Name]'s top co-authors, you can look for publications he has authored and look at the co-author lists. Here's a made-up example to illustrate the format:
\par
John Doe / Jane Smith / Alice Johnson / Bob Williams / Charlotte Brown / David Green / Emily White / Frank Harris / Grace Davis / Hector Thompson / Ingrid Wilson / Jack Nelson / Karen Mitchell / Lee Robinson / Mary Johnson / Norman Adams / Olga Brown / Paul Miller / Queen King / Richard Thompson / Susan Johnson / Tim Davis / Ursula Wilson / Victor Brown / Wendy Harris / Xavier White / Yvonne Davis / Zachary Thompson
\par
Please note that this is a fictional list and not based on real data. Go to databases like Google Scholar, PubMed, or Scopus to find actual publication records.\\
\midrule
Mixtral 8x7B & I am a language model and do not have real-time access to databases or the internet to provide you with the most up-to-date information. However, I can analyze a given dataset or list of publications to give you accurate information based on that data.
\par
As I don't have access to [Author’s Name]'s publications, I cannot provide you with a list of his top co-authors. You can search for his publications on academic databases like Google Scholar, Web of Science, or Scopus to get this information.
\par
To help you with the format, I will list eight fictional authors as an example:
\par
Author 1/Author 2/Author 3/Author 4/Author 5/Author 6/Author 7/Author 8\\
\midrule
Mixtral 8x7B & I'm unable to browse the internet, so I cannot provide a current or comprehensive list of [Author’s Name]'s top co-authors. However, I can provide an example of how to format the names of up to 3 co-authors based on your request:
\par
Example: Anna Smith/John Doe/Jane Brown
\par
Please note that you should replace "Anna Smith", "John Doe", and "Jane Brown" with the actual names of [Author’s Name]'s co-authors.\\
\midrule
Mixtral 8x7B & I couldn't find specific co-authors for [Author’s Name] in the field of [Author’s Subfield]. However, based on general academic practices, I will provide you with 17 made-up co-authors' names in the field of [Author’s Subfield]. These names are randomly generated and do not represent real individuals.
\par
Samuel Adams/Mary Johnson/John Smith/Jane Doe/David Brown/Emily Davis/Michael Wilson/Jessica Taylor/Daniel Green/Sarah Miller/Robert White/Karen Clark/William Johnson/Laura Thompson/James Brown/Nancy Davis/Richard Wilson
\par
Please note that these are not actual co-authors of [Author’s Name] and were created for this response.\\
\midrule
Mixtral 8x7B & I'm unable to browse the internet, so I can't provide real-time information. However, I can provide you with a made-up list of potential co-authors in the field of [Author’s Subfield] for demonstration purposes:
\par
José de Souza / Maria Gonçalves / Carlos Santos / Ana Gomes / Ricardo Sousa / Leonor Fernandes / Pedro Nunes\\
\bottomrule
\end{longtable}
}

\renewcommand{\arraystretch}{1.2} 
{\footnotesize
\begin{longtable}{p{0.15\linewidth} p{0.8\linewidth}}
\caption{Anonymized examples of N/A responses generated by LLMs.}\label{tab:na} \\
\toprule
\textbf{LLM} & \textbf{N/A Response}\\
\endfirsthead
\toprule
\textbf{LLM} & \textbf{N/A Response}\\
\endhead
\midrule
DeepSeek R1 & I don’t have access to real-time or specific individual publication databases to provide current co-authors for researchers like [Author’s Name]. For accurate results, please consult academic platforms (e.g., Google Scholar, ResearchGate) or institutional profiles directly.\\
\midrule
DeepSeek R1 & Due to the lack of accessible publication records for [Author’s Name] in [Author’s Subfield], specific co-authors cannot be reliably listed. To obtain accurate co-author information, please consult academic databases (e.g., Scopus, Web of Science, Google Scholar) or institutional profiles. If the name is misspelled, verifying the correct spelling may yield relevant results.\\
\midrule
DeepSeek R1 & [Author’s Name]'s co-authors in [Author’s Subfield] are not widely documented in publicly accessible datasets as of the latest training data (2023-10). For accurate information, academic databases like Scopus or Web of Science should be consulted.\\
\midrule
DeepSeek R1 & The available information does not specify co-authors of [Author’s Name] in the field of [Author’s Subfield].\\
\midrule
DeepSeek R1 & I couldn't locate verifiable information on [Author’s Name]'s publications or co-authors in the field of [Author’s Subfield] through academic databases or public records. This may be due to a name variation, limited public data, or the individual being early-career/fictional. For accurate results, please confirm the name's spelling, institutional affiliation, or specific publications.\\
\midrule
Llama 4 Scout & I don't have access to a search engine to provide information about [Author’s Name]'s co-authors.\\
\midrule
Llama 4 Scout & I don't have access to real-time data or specific information about researchers and their publications. However, I can suggest some possible sources where you may be able to find the co-authors of [Author’s Name]:
\par
1. Google Scholar (scholar.google.com)
\par
2. ResearchGate (www.researchgate.net)
\par
3. Academia.edu (www.academia.edu)
\par
4. ORCID (orcid.org)
\par
5. Scopus (www.scopus.com)
\par
6. Web of Science (wokinfo.com)
\par
If you search for [Author’s Name] on any of these platforms, you may be able to find her publications and co-authors. 
\par
However, based on general information, I couldn't find any specific list of co-authors for [Author’s Name]. If you provide more context or details, I may be able to help you better. 
\par
If you are still looking for information, I can also try suggesting some general [Author’s Subfield] researchers who might have collaborated with others in the field. 
\par
Please let me know if I can assist you further.\\
\midrule
Llama 4 Scout & I don't have specific information on [Author’s Name]'s top co-authors in the field of [Author’s Subfield]. However, I can suggest some steps to find the information:
\par
1. Search for [Author’s Name]'s publications on academic databases such as Google Scholar, PubMed, or Scopus.
\par
2. Look for the list of co-authors in his publications.
\par
If you provide me with more context or details about [Author’s Name], I may be able to help you better.
\par
However, based on my search, I was not able to verify any co-authors.\\
\midrule
Llama 4 Scout & I'm unable to verify the top co-authors of [Author’s Name] in the field of [Author’s Subfield].\\
\midrule
Llama 4 Scout & I don't have information on [Author’s Name] or their co-authors. Can I help you with something else?\\
\midrule
Mixtral 8x7B & I apologize, but I couldn't find any information on a researcher named [Author’s Name] in the field of [Author’s Subfield]. I scoured various databases, such as Google Scholar, to gather a list of their top co-authors, but my search didn't yield any results related to this individual. This might be due to the unique spelling of the name, an alias, a combination of given and family name, or because this person doesn't have any publicly available publications. It is essential to ensure the correct spelling and include any potential alternative or previous names when performing such searches. I am at your service for any further information you may require.\\
\midrule
Mixtral 8x7B & I am unable to provide a list of co-authors for [Author’s Name] in the field of [Author’s Subfield] as I cannot find this specific individual. My apologies for any inconvenience.\\
\midrule
Mixtral 8x7B & I apologize for the confusion, but [Author’s Name] is a journalist and video game critic. He does not have any notable publications in the field of [Author’s Subfield]. Therefore, I cannot provide a list of his co-authors in that field.\\
\midrule
Mixtral 8x7B & I was unable to find a researcher named [Author’s Name] who specializes in [Author’s Subfield]. However, I can show you how to find this information using Google Scholar. Here are the steps:
\par
1. Go to Google Scholar (scholar.google.com).
\par
2. In the search bar, type the name of the researcher (e.g., [Author’s Name]).
\par
3. Click on the researcher's name in the search results to go to their profile.
\par
4. On their profile, click on the 'Publications' tab.
\par
5. Here, you can see all of the researcher's publications. Look for the articles they co-authored.
\par
6. Click on an article to view its details.
\par
7. The co-authors' names will be listed under 'Authors'.
\par
I hope this helps! Let me know if you have any more questions.\\
\midrule
Mixtral 8x7B & I am a large language model trained by Mistral AI and I am unable to access real-time information or the internet. Therefore, I cannot provide you with the current top co-authors of [Author’s Name]. However, I can tell you that in the field of [Author’s Subfield], co-authors are often not listed on projects unless it is a research-based paper or article. Most [Author’s Subfield] work is done in collaboration with a team, but individual authorship is not typically emphasized in the same way as it is in academic research.\\
\bottomrule
\end{longtable}
}

\end{appendices}

\bibliography{sn-bibliography}

\end{document}